# GRPO and Reflection Reward for Mathematical Reasoning in Large Language Models

Zhijie Wang

*School of Computer Science and Engineering, Sun Yat-sen University(SYSU), Guangzhou, China*
*wangzhj79@mail2.sysu.edu.cn*

*Abstract:* The enhancement of reasoning capabilities in large language models (LLMs) has garnered significant attention, with supervised fine-tuning (SFT) and reinforcement learning emerging as dominant paradigms. While recent studies recognize the importance of reflection in reasoning processes, existing methodologies seldom address proactive reflection encouragement during training. This study focuses on mathematical reasoning by proposing a four-stage framework integrating Group Relative Policy Optimization (GRPO) with the reflection reward mechanisms to strengthen LLMs' self-reflective capabilities. Besides, this approach incorporates established accuracy and format reward. Experimental results demonstrate GRPO's state-of-the-art performance through reflection-encouraged training, with ablation studies confirming the reflection reward's pivotal role. Comparative evaluations demonstrate full-parameter SFT's superiority over low-rank adaptation (LoRA) despite heightened computational demands. Building on these cumulative findings, this research substantiates GRPO's methodological significance in post-training optimization and envisions its potential to serve as a pivotal enabler for future LLM-based intelligent agents through the synergistic integration of cognitive rewards with dynamic environmental interactions.

## 1. Introduction

The emergence of domain-specific Artificial Intelligence has catalyzed a paradigm shift toward specialized models as critical computational infrastructure. This paradigm shift necessitates cost-efficient adaptation strategies, with post-training methods establishing dominance due to the prohibitive computational expenditures required for training from scratch.

As summarized by Kumar et al., both fine-tuning and reinforcement learning (RL) methodologies have demonstrated significant potential in advancing the reasoning capabilities of large language models (LLMs) [1]. Empirical evidence from Google Research reveals a complementary relationship: fine-tuning primarily enhances knowledge retention, whereas RL-driven approaches exhibit superior generalization performance [2].

Within this landscape, Group Relative Policy Optimization (GRPO) has emerged as a methodologically significant framework, particularly evidenced by its state-of-the-art performance in the DeepSeek R1 training process [3, 4]. Concurrently, supervised fine-tuning (SFT) and low-rank adaptation (LoRA) represent distinct optimization paradigms – SFT excels in task-specific alignment through direct supervision, while LoRA achieves parameter-efficient adaptation via low-rank matrix decomposition [5].

A critical insight from the DeepSeek R1 technical report highlights that language models exhibit "aha moments", where reevaluation of initial reasoning trajectories triggers nonlinear performance improvements [4]. Inspired by this phenomenon, this study introduces a novel reflection reward mechanism. This component is integrated into the GRPO training framework to enhance LLMs' mathematical reasoning capabilities through self-reflective learning. Extending the Open R1 project, this research achieves three primary objectives: (1) development of an enhanced language model demonstrating robust mathematical reasoning capabilities, (2) innovative integration of reflection reward with ablation studies to validate its effectiveness, and (3) systematic comparison of fine-tuning methodologies [6]. These contributions collectively advance the understanding of LLM reasoning optimization.

The code and experimental results have been uploaded to a GitHub repository developed by the author [7].

## 2. Methodologies

### 2.1. Datasets

This study adopts a task-segmented training framework with specialized dataset allocation for different training phases. Specifically, during the fine-tuning phase, the framework leverages openr1/OpenR1-Math-220k dataset, a curated corpus of 220,000 mathematical problems [8]. This dataset is constructed by the Open R1 research team via DeepSeek-R1's trace generation from 400,000 NuminaMath 1.5 base problems. For the GRPO phase, the DigitalLearningGmbH/MATH-lighteval dataset provides mathematical problems at competition level, with solutions available in both LaTeX-formatted proofs and natural language explanations [9].

The evaluation datasets incorporate three specialized benchmarks: (1) AIME2024 evaluates mathematical reasoning at the high school competition level [10]. (2) MATH-500 verifies multi-step solution integrity via 500 curated, proof-based problems [11]. (3) GPQA: Diamond evaluates cross-disciplinary analytical proficiency using 198 graduate-level STEM questions requiring synthetic knowledge integration [12].

### 2.2. Workflow

The architecture implements a framework through four progressive stages. First, the Base Model Deployment Stage initializes the pre-trained model. Subsequently, the Fine-tuning Stage employs SFT and LoRA for domain-specific knowledge injection. During the GRPO Stage, group-relative reward estimation integrates three core metrics: accuracy, format, and reflection. Finally, the Testing & Evaluation Stage adopts accuracy metrics on three standardized benchmark datasets (aime24, math-500, and gpqa: diamond).

#### 2.2.1. Base model deployment

During the Base Model Deployment Stage, with comprehensive consideration of computational resource constraints, this study adopts Qwen2.5-Math as the base model instead of training a language model from scratch [13]. Through evaluation on the MATH-500 benchmark, the 1.5B parameter variant achieves an accuracy rate of 45.4%, while the 7B variant attains 58.6% accuracy. These metrics serve as the baseline for subsequent comparative analysis with reasoning models.

### 2.2.2. Fine-Tuning

Following the benchmarking of the base model, the research proceeds to the Fine-tuning Stage, employing the OpenR1-Math-220k mathematical reasoning dataset generated by DeepSeek R1 for task-specific adaptation. This study systematically compares the technical characteristics and empirical results of three strategies: SFT, LoRA, and a non-fine-tuned baseline.

SFT, a full parameter fine-tuning approach, conducts comprehensive parameter adjustment across all network layers to align model outputs through supervised training. As demonstrated by Raffel et al., this approach enables domain adaptation by exhaustively exploring the parameter space of pre-trained architectures, though requiring considerable computational resource demands that necessitate specialized hardware infrastructure [14].

In contrast, LoRA, a parameter-efficient fine-tuning technique, implements parameter-efficient adaptation through low-rank decomposition of weight updates. Specifically, it decomposes weight matrix increments $\Delta W$ into the product of two low-rank matrices $B$ and $A^T$:

$$\Delta W = BA^\top (B \in R^{d \times r}, A \in R^{k \times r}) \tag{1}$$

Where the rank hyperparameter r is substantially smaller than matrix dimensions $d$ and $k$, thereby compressing trainable parameters from $d \times k$ to $r(d+k)$. Experimental results from Dettmers et al. reveal that LoRA substantially reduces the memory overhead during parameter adaptation, while preserving model performance comparable to the original [15].

### 2.2.3. GRPO

Furthermore, the workflow progresses to the GRPO Stage, where this study implements a comparative experimental design to evaluate the optimization efficacy of two initialization strategies: (1) SFT variant, wherein policy optimization commences from a fine-tuned model; (2) None fine-tuned variant, employing the original parameters of the pretrained base model for policy initialization. To conduct an ablation study with the results reproduced by the OpenR1 research team based on the study of Zeng et al., the training dataset is deliberately decoupled from the SFT data by substituting the MATH-lighteval mathematical reasoning dataset [16].

The GRPO optimization initiates with group sampling to generate multiple candidate response sequences. A tripartite reward function subsequently calculates normalized relative advantages: Accuracy Reward quantifies final answer correctness, Format Reward enforces structured output specifications, and the innovative Reflection Reward drives introspective evaluation of reasoning trajectories. The concept of reflection reward is inspired by the "aha moments" described in the DeepSeek R1 technical report, which highlights that language models can undergo sudden insights, where reassessing initial reasoning pathways leads to nonlinear performance improvements [4]. Additionally, this idea is influenced by prior studies, such as the Self-RAG framework, which enhances text generation quality through retrieval-augmented self-reflection [17]. Moreover, the work of Ding et al. introduces a cooperative learning paradigm where a smaller model acts as a study assistant to facilitate the reflection process of a larger model [18]. Specifically, the design of the reflection reward is informed by the work of Gandhi et al., structured along four key cognitive dimensions: verification, backtracking, subgoal setting, and backward chaining [19]. In addition, diverging from conventional Proximal Policy Optimization (PPO) paradigms, this algorithm eliminates the critic model to reduce computational overhead. Moreover, to curb policy drift, GRPO integrates KL-divergence regularization within the objective function:

$$J_{\text{GRPO}}(\theta) = \frac{1}{G}\sum_{i=1}^{G} E_{(x,a_i)\sim\pi_{\theta_{\text{old}}}}\left[\min\left(\frac{\pi_\theta(a_i|x)}{\pi_{\theta_{\text{old}}}(a_i|x)}\tilde{r}_i, \text{clip}\left(\frac{\pi_\theta(a_i|x)}{\pi_{\theta_{\text{old}}}(a_i|x)}, 1-\epsilon, 1+\epsilon\right)\tilde{r}_i\right)\right] - \beta D_{\text{KL}}\left(\pi_\theta|\pi_{\text{ref}}\right) \quad (2)$$

In the GRPO objective function above, $\theta$ denotes the current policy parameters, and $G$ is the number of candidate responses generated via group sampling. For each candidate indexed by $i$, $x$ represents the input context, and $a_i$ is the corresponding action sampled from the previous policy $\pi_{\theta_{\text{old}}}$. The ratio $\frac{\pi_\theta(a_i|x)}{\pi_{\theta_{\text{old}}}(a_i|x)}$ reweights the normalized advantage $\tilde{r}_i$, which is computed as a linear combination of Accuracy, Format, and Reflection rewards, with the weights regulated by the hyperparameter λ, and defined by the formula:

$$\tilde{r}_i = \lambda_{\text{acc}} R_{\text{acc}} + \lambda_{\text{fmt}} R_{\text{fmt}} + \lambda_{\text{refl}} R_{\text{refl}} \quad (3)$$

Additionally, the clipping function $\text{clip}(\cdot, 1-\epsilon, 1+\epsilon)$ constrains the update magnitude with $\epsilon$ as a small positive constant, while the KL divergence term $D_{\text{KL}}(\pi_\theta|\pi_{\text{ref}})$, scaled by β, regularizes the policy update by penalizing deviation from a reference policy $\pi_{\text{ref}}$.

### 2.2.4. Testing & evaluation

In the final stage of the workflow, this study implements a multidimensional evaluation framework to systematically quantify reasoning capabilities through accuracy metrics, contrasting with the Qwen2.5-Math baseline across three mathematical reasoning benchmarks: MATH-500, GPQA: Diamond, and AIME24.

## 3. Results

As demonstrated in Table 1, the comprehensive evaluation reveals that the Qwen-2.5-Math-7B-Reasoning (GRPO-only) model, trained on the MATH-lighteval dataset, exhibits superior performance characteristics. Specifically, by implementing a composite reward mechanism integrating three components - reflection reward, accuracy reward, and format reward - the model achieves an accuracy of 73.8% on the MATH-500 evaluation set. This improvement represents a 15.2% absolute increase in accuracy over the baseline Qwen2.5-Math-7B model, thereby validating the efficacy of the GRPO in augmenting LLM reasoning capabilities.

Table 1: Quantitative evaluation of reasoning capabilities (table credit: original)

| Model | MATH-500 | gpqa: diamond | AIME_2024 |
|---|---|---|---|
| Qwen2.5-Math-1.5B | 45.4 | 27.27 | 10.00 |
| Qwen2.5-Math-7B | 58.6 | 26.26 | 16.67 |
| Qwen2.5-Math-1.5B-sft | 58.2 | 35.35 | 3.33 |
| Qwen2.5-Math-1.5B-LoRA-Merged | 42.2 | 30.81 | 6.67 |
| Qwen-2.5-Math-1.5B-Reasoning(sft+grpo) | 50.0 | 26.26 | 13.33 |
| Qwen-2.5-Math-1.5B-Reasoning(only grpo) | 54.6 | 25.76 | 13.33 |
| Qwen2.5-Math-7B-sft | 66.2 | 37.37 | 16.67 |
| Qwen-2.5-Math-7B-Reasoning(sft+grpo) | 57.6 | 41.41 | 16.67 |
| **Qwen-2.5-Math-7B-Reasoning(only grpo)** | **73.8** | **34.34** | **20.00** |

Notably, ablation studies conducted through comparative analysis with openr1's technical report demonstrate the significance of the reflection reward. According to theOpen R1 research team, constraining the GRPO framework through the exclusion of reflection rewards, while maintaining accuracy and format reward mechanisms, leads to a performance reduction of 4.4 percentage points (69.4% vs. 73.8%) [6]. In addition, as depicted in Figure 1, both the red curve (representing the Qwen-2.5-7B-Reasoning model) and the green curve (representing the Qwen-2.5-1.5B-Reasoning model) exhibit a steadily increasing reflection reward. Moreover, the larger-parameter model (red curve) demonstrates a more pronounced upward trend, thereby yielding more substantial performance gains overall. These evidences strongly suggest that the reflection reward mechanism is essential in iterative reasoning refinement throughout the model's training process.

Furthermore, a systematic comparison is undertaken between full-parameter fine-tuning and parameter-efficient adaptation approaches. LoRA-merged models exhibit a 3.2% accuracy reduction on MATH-500 relative to the baseline, potentially attributable to a mismatch between pre-training and downstream data distributions. This pronounced sensitivity presents a contrast to SFT models, which maintain consistent performance over LoRA variants, validating the efficacy of full-parameter optimization despite incurring higher computational overhead.

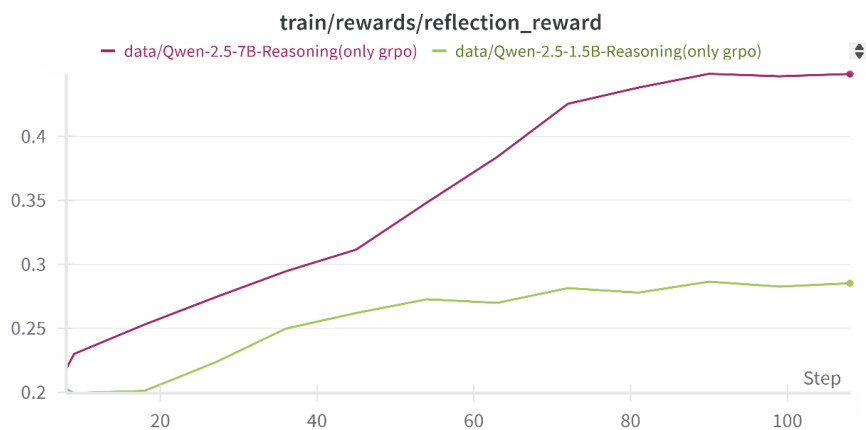

Figure 1: Temporal dynamical improvement of reflection reward during GRPO training phase (picture credit: original)

## 4. Limitations and perspectives for the future

However, computational resource constraints preclude comprehensive experimentation with large-scale datasets. As evidenced in Table 1, the GRPO-trained models based on SFT fine-tuning underperform, primarily due to the significantly smaller dataset used in GRPO compared to SFT, leading to significant distributional discrepancies between their respective training data. Additionally, the current investigation remains confined to specialized mathematical reasoning tasks, leaving broader explorations of cross-task generalization capabilities for future research.

Future research should prioritize three key enhancements. First, a systematic refinement of the SFT and GRPO datasets is recommended, as their synergistic integration with the reflection reward mechanism may significantly amplify reflective capabilities. Second, further optimization of system prompts and GRPO reward strategies is essential to strengthen structured reasoning. This can be achieved by constraining generation modes—for instance, by introducing specific tags for iterative verification (like <rethink>) or external resource invocation (like <tool>). Third, extending this paradigm to incorporate dynamic interactions with external environments could foster the development of tool-usage competencies, thereby paving the way for advanced LLM agent systems.

## 5. Conclusion

This study establishes a four-stages progressive framework demonstrating GRPO's efficacy in mathematical reasoning enhancement. Empirical results show that GRPO with reflection reward achieves 15.2% absolute accuracy improvement (73.8% vs. 58.6%) on MATH-500. Notably, ablation analysis confirms the critical role of reflection rewards, whose removal causes 4.4% performance degradation. Furthermore, comparative evaluations substantiate full-parameter fine-tuning's domain adaptation superiority over LoRA (+16% accuracy).

In conclusion, GRPO effectively enhances large language model reasoning, establishing its methodological significance in post-training paradigms. Building on this methodological foundation, future investigations necessitate development of dynamic multi-objective reward calibration and anthropocentric reflection frameworks that maintain cognitive alignment with human reasoning patterns.